\documentclass[conference]{IEEEtran}
\IEEEoverridecommandlockouts
\usepackage{cite}
\usepackage{amsmath,amssymb,amsfonts}
\usepackage{algorithm}
\usepackage{algorithmic}
\usepackage{graphicx}
\usepackage{textcomp}
\usepackage{booktabs} 
\usepackage{xcolor}
\usepackage{hyperref}
\usepackage{float} % Make sure this is in your document preamble
\def\BibTeX{{\rm B\kern-.05em{\sc i\kern-.025em b}\kern-.08em
    T\kern-.1667em\lower.7ex\hbox{E}\kern-.125emX}}

\begin{document}

\title{Exploration-Driven Personalized Federated Reinforcement Learning via Intrinsic Motivation}

\author{
\IEEEauthorblockN{Md Rafid Islam}
\IEEEauthorblockA{Department of Electrical and Computer Engineering\\
North South University\\
Dhaka, Bangladesh\\
md.islam.241@northsouth.edu}
\and
\IEEEauthorblockN{Rafsan Jany}
\IEEEauthorblockA{Digital Health Research Division\\
Korea Institute of Oriental Medicine\\
Daejeon, South Korea\\
rafsanjany@kiom.re.kr}
\and
\IEEEauthorblockN{Zahid Hasan}
\IEEEauthorblockA{Department of Electrical and Computer Engineering\\
North South University\\
Dhaka, Bangladesh\\
zahid.hasan.241@northsouth.edu}
\and
\IEEEauthorblockN{Ratun Rahman}
\IEEEauthorblockA{Department of Electrical and Computer Engineering\\
The University of Alabama in Huntsville\\
Huntsville, AL, USA\\
rr0110@uah.edu}
}

\maketitle

\begin{abstract}
Personalized Federated Reinforcement Learning (PFRL) takes a decentralized approach to storing and accessing information based on past experiences while keeping each client's data private during the learning of each client's policy. Many current methods for PFRL rely heavily on exploiting existing reinforcement learning reward signals to derive an optimal policy for each client, thereby neglecting exploration in non-stationary or sparse-reward environments. In this work, we introduce a new exploration-driven framework, Exploration-Driven Personalized Federated Reinforcement Learning via Intrinsic Motivation (EDPFRL-IM), that leverages an inherent curiosity-driven exploration at each client to promote local exploration and protect client privacy. Furthermore, to facilitate policy discovery via exploration in previously unexplored state spaces, clients add an intrinsic random network distillation (RND) signal to their extrinsic reward. Additionally, the server does not have access to clients' raw experiences or local gradient estimates; instead, the server sends global exploration priors and collects minimal novelty summaries from each client to enable both diverse and coordinated exploration among clients. Experiments in benchmark environments show that our framework outperforms average PFRL benchmarks in policy personalization and sample efficiency, primarily in delayed and sparse reward systems. Overall, EDPFRL-IM enables the integration of a flexible exploratory learning structure into federated reinforcement learning systems while preserving client privacy.
\end{abstract}
 
 
\section{Introduction}
Reinforcement learning (RL) is a powerful framework for sequential decision-making that enables agents to learn optimal behaviors through interaction with their environment~\cite{shakya2023reinforcement}. Federated reinforcement learning (FRL) enables multiple clients to cooperatively train reinforcement learning (RL) agents in decentralized settings without exchanging raw experience data. Personalized FRL (PFRL) expands on this by customizing policies to each client's particular environment and objectives. This makes it suitable for real-world applications such as health monitoring and robotics~\cite{fan2021fault, rahman2024improved}.
 
Although existing PFRL frameworks~\cite{fan2021fault, wang2020optimizing} focus on policy optimization and aggregation, they often overlook the role of exploration under limited or delayed rewards. Standard RL strategies~\cite{shakya2023reinforcement} like $\epsilon$-greedy or entropy regularization become less effective in federated settings, where clients operate independently and face privacy constraints that limit coordination. As a result, clients may learn suboptimal policies, particularly in cold-start or non-stationary conditions. While approaches like FedRL~\cite{fan2021fault}, FedAvg-RL~\cite{wang2020optimizing}, and pFedMe~\cite{t2020personalized} address personalization or aggregation, they assume homogeneous or supervised settings. Intrinsic motivation methods such as RND~\cite{nikulin2023anti}, ICM~\cite{colas2019curious}, and count-based exploration~\cite{machado2020count} show promise for centralized RL but remain underexplored in federated scenarios. Addressing this gap is key to unlocking more adaptive, intelligent on-device learning.
 
We propose the \textit{EDPFRL-IM} framework, which integrates intrinsic curiosity-driven motivation into personalized federated reinforcement learning. It not only enables efficient local exploration with sparse rewards but also preserves privacy and minimizes communication. We summarize our key contributions as follows.
 
\begin{itemize}
    \item We introduce \textit{EDPFRL-IM}, a novel FRL framework that incorporates intrinsic motivation for personalized exploration in a decentralized setting.
    \item We develop a communication-efficient protocol for compressing exploration statistics, enabling clients to align their curiosity-driven behavior while avoiding disclosure of raw data or policies.
    \item We empirically evaluate our approach in sparse-reward benchmark environments and demonstrate significant improvements in personalization, exploration efficiency, and policy performance over existing federated RL baselines.
\end{itemize}
 
\section{Methodology}
 
\subsection{Problem Formulation}
 
We consider a PFRL setting with $N$ clients, each interacting with a local Markov decision process (MDP) \cite{lauri2022partially} $\mathcal{M}_i = (\mathcal{S}_i, \mathcal{A}, \mathcal{P}_i, R_i, \gamma)$, where $\mathcal{S}_i$ and $R_i$ are client-specific, while $\mathcal{A}$ and $\gamma$ are shared. Due to heterogeneity in transition dynamics and rewards, each client learns a personalized policy $\pi_i(a|s)$ to maximize its expected return:
\begin{equation}
J_i(\pi_i) = \mathbb{E}_{\pi_i}\left[ \sum_{t=0}^{\infty} \gamma^t r_{i,t} \right].
\end{equation}
Clients update their policy parameters via gradient ascent:
\begin{equation}
\theta_i \leftarrow \theta_i + \eta \nabla_{\theta_i} J_i(\pi_{\theta_i}),
\end{equation}
with learning rate $\eta$. Unlike global FL or RL, PFRL optimizes individual client policies while enabling collaboration through lightweight, privacy-preserving exploration statistics.

\subsection{Intrinsic Motivation for Local Exploration}
Effective exploration is crucial in PFRL due to sparse or delayed rewards. We adopt RND, a self-supervised method in which each client uses a fixed random target network $f_{\text{tgt}}$ and a trainable predictor $f_{\text{pred}}$. The intrinsic reward is the prediction error on the next state $s_{t+1}$:
\begin{equation}
r_t^{\text{int}} = \left\| f_{\text{tgt}}(s_{t+1}) - f_{\text{pred}}(s_{t+1}) \right\|_2^2.
\end{equation}
 
This promotes the exploration of novel states. The total reward combines extrinsic and intrinsic terms as
\begin{equation}
r_t^{\text{total}} = r_t^{\text{ext}} + \alpha_i \cdot r_t^{\text{int}},
\end{equation}
where $\alpha_i$ controls the exploration-exploitation trade-off per client. This local RND setup enables adaptive exploration without external shaping or manual tuning.
 
\subsection{Federated Coordination with Exploration Summaries}
Uncoordinated exploration can lead to redundant efforts across clients. To enable efficient collaboration, we propose a federated privacy-preserving coordination mechanism. Each client generates compact exploration statistics $\mathcal{E}_i$, such as visitation counts, frequency histograms over state embeddings, or top-$k$ novel state hashes based on intrinsic rewards. These summaries are low-dimensional and protect sensitive data.
 
The central server aggregates these statistics to form a global exploration profile as
\begin{equation}
\mathcal{G} = \text{Aggregate}(\{ \mathcal{E}_i \}_{i=1}^N),
\end{equation}
where $\text{Aggregate}(\cdot)$ combines client contributions, e.g., summing visitation counts for the state cluster $k$ as
\begin{equation}
v_{\text{global}}[k] = \sum_{i=1}^N v_i[k].
\end{equation}
 
The server then computes a global novelty prior to identifying underexplored states:
\begin{equation}
\mathcal{P}_{\text{novel}}(s) = \frac{1}{1 + v_{\text{global}}[\text{cluster}(s)]},
\end{equation}
where $\text{cluster}(s)$ maps the state $s$ to a cluster index. This prior is broadcast to all clients to guide their exploration.
 
Each client incorporates the global novelty prior into its policy update as
\begin{equation}
\theta_i \leftarrow \theta_i + \eta \nabla_{\theta_i} \mathbb{E}_{\pi_{\theta_i}, p_i(s)} \left[ \sum_{t=0}^{\infty} \gamma^t (r_{i,t}^{\text{ext}} + \alpha_i r_{i,t}^{\text{int}}) \right],
\end{equation}
where the probability of sampling state $s$ from the experience buffer of client $i$, $\mathcal{D}_i$, is expressed as
\begin{equation}
p_i(s) \propto \left(1 + \beta \cdot \mathcal{P}_{\text{novel}}(s)\right) \cdot p^{\text{uniform}}_i(s),
\end{equation}
with $p^{\text{uniform}}_i(s) = 1/|\mathcal{D}_i|$ as the uniform baseline probability and $\beta \geq 0$ controlling the influence of the global prior. This approach ensures that clients focus on globally novel states, promoting diverse exploration while maintaining privacy and low communication costs.
 
\begin{figure*}[t]
    \centering
    \includegraphics[width=0.85\textwidth]{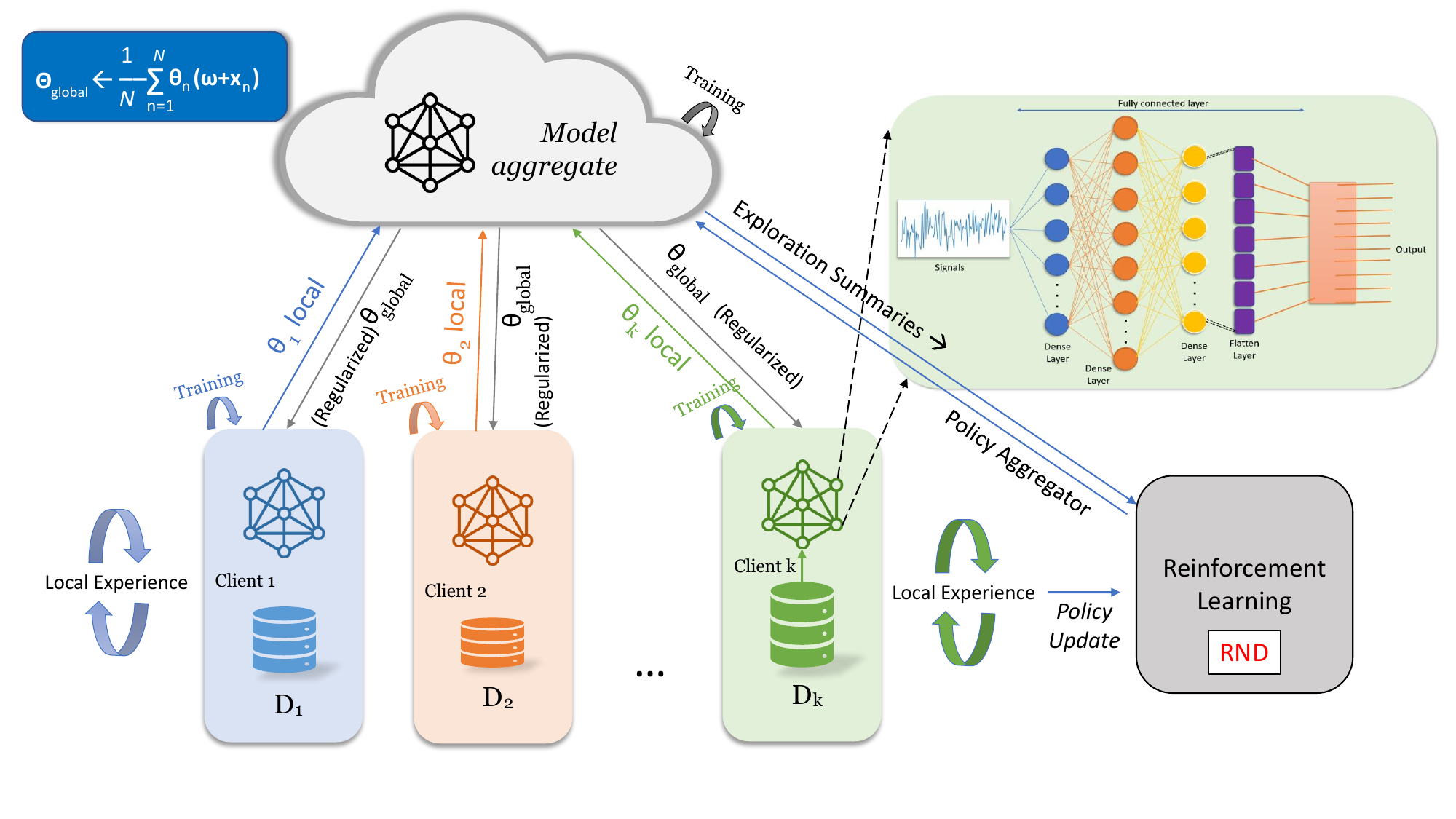}
    \caption{An overview of the proposed EDPFRL-IM framework. Based on local experience, each client uses intrinsic incentives to do reinforcement learning (via RND). Clients provide compressed exploration summaries to a central server regularly, which are then aggregated into a global novelty prior. This prior is transmitted back to guide tailored exploration, allowing clients to explore together while maintaining their private information.}
    \vspace{-1em}
    \label{fig:placeholder}
\end{figure*}

Algorithm~\ref{alg:edpfl} describes the EDPFRL-IM method. Each client gathers experience and computes intrinsic rewards using RND (Lines 5--7), then adjusts its local policy (Line 8). Clients transmit compressed exploration summaries to the server (Line 10), which aggregates them into a global novelty prior (Line 12) and broadcasts it (Line 13). Clients use this prior for biased experience sampling (Line 15), which allows for coordinated exploration without compromising privacy. An overview of our framework is shown in Fig.~\ref{fig:placeholder}.
 
\begin{algorithm}[t]
\footnotesize
\caption{\footnotesize EDPFRL-IM: Exploration-Driven Personalized Federated Reinforcement Learning}
\label{alg:edpfl}
\begin{algorithmic}[1]
\REQUIRE Clients $\{C_i\}_{i=1}^{N}$, global rounds $T$, local epochs $E$, RND module $(f_{\text{tgt}}, f_{\text{pred}})$, exploration weight $\alpha_i$, novelty bias $\beta$
 
\FOR{each global round $t = 1$ to $T$}
    \FOR{each client $C_i$ in parallel}
        \STATE Initialize local buffer $\mathcal{D}_i$
        \FOR{local step $e = 1$ to $E$}
            \STATE Interact with environment to collect $(s_t, a_t, r_t^{\text{ext}}, s_{t+1})$
            \STATE Compute intrinsic reward: $r_t^{\text{int}} = \left\| f_{\text{tgt}}(s_{t+1}) - f_{\text{pred}}(s_{t+1}) \right\|_2^2$
            \STATE Store $(s_t, a_t, r_t^{\text{total}}, s_{t+1})$ in $\mathcal{D}_i$, where $r_t^{\text{total}} = r_t^{\text{ext}} + \alpha_i \cdot r_t^{\text{int}}$
            \STATE Update policy $\pi_i$ using local RL optimizer (e.g., PPO, SAC)
        \ENDFOR
        \STATE Compute exploration summary (e.g., histogram, top-$k$ novel states)
        \STATE Send summary to server
    \ENDFOR
    \STATE Server aggregates summaries into global novelty prior $\mathcal{P}_{\text{novel}}$
    \STATE Broadcast $\mathcal{P}_{\text{novel}}$ to all clients
    \FOR{each client $C_i$}
        \STATE Update sampling strategy: \\
        \quad $p_i(s) \propto (1 + \beta \cdot \mathcal{P}_{\text{novel}}(s)) \cdot p^{\text{uniform}}_i(s)$
    \ENDFOR
\ENDFOR
\end{algorithmic}
\end{algorithm}
 
\section{Experiments}
 
\subsection{Simulation Setup}
 
\textbf{Environments:} We test the performance of EDPFRL-IM in standard reinforcement learning environments with sparse rewards and heterogeneous dynamics. We employ modified versions of \texttt{MountainCar-v0} \cite{chavali2022modelling} and \texttt{CartPole-sparse} \cite{bjorck2021towards}, which are chosen for their sensitivity to exploration quality and their significance in comparing sample efficiency.
 
\textbf{Federated Setup:} We simulate a realistic FRL scenario by considering $N = 10$ clients, each running in a different local environment built from \texttt{MountainCar-v0} and a modified \texttt{CartPole} environment with sparse rewards. We create heterogeneity among clients by adjusting the elements of the environment, including gravity, friction, and reward shaping. For instance, in \texttt{MountainCar-v0}, clients earn additional shaping incentives based on distance or velocity, and gravity varies from $0.0025$ to $0.006$. In \texttt{CartPole-sparse}, clients vary in pole mass and cart friction, and earn a reward only after balancing for a minimum duration. As an example, one client's \texttt{MountainCar-v0} variation employs gravity $0.0045$, friction $0.005$, and a shaped reward of $r_t = -1 + 0.1 \cdot |x - x_{\text{goal}}|$ to promote goal-directed travel. These differences lead to different MDPs, indicating extremely non-IID conditions where effective policy learning requires personalization.
 
\textbf{Baseline Comparison:} Training proceeds for $T = 100$ global communication rounds, with each client performing $E = 10$ local PPO updates per round. Policies are represented by fully connected two-layer neural networks with 64 hidden units and ReLU activations. Each client has its own Random Network Distillation (RND) module, with both the target and predictor networks consisting of a single hidden layer of size 128. The intrinsic reward coefficient $\alpha_i$ is set to 0.1, while the novelty sampling bias $\beta$ is set to 0.5, unless specified otherwise.
 
\textbf{Implementation Details:} We compare our approach with the following baselines: (i) \textit{Local RL}, in which every client trains on its own without any federated coordination; (ii) \textit{FedRL}, a common FRL method with global policy aggregation; and (iii) \textit{FedRL+RND}, which applies intrinsic rewards locally without exploration coordination. For the broader comparison in Table~\ref{tab:final_comparison}, we additionally report single-agent intrinsic-motivation methods run in the non-federated setting (RND~\cite{nikulin2023anti}, ICM~\cite{colas2019curious}, and count-based exploration (CBE)~\cite{machado2020count}), as well as federated and personalized FL baselines without coordinated exploration (FedAvg-RL~\cite{wang2020optimizing}, pFedMe~\cite{t2020personalized}, and FedPer++~\cite{xu2022fedper++}). All baselines are trained under the same environment configurations, client counts, and total communication budget as EDPFRL-IM for a fair comparison. The Adam optimizer is used to train all models, with a discount factor of $\gamma = 0.99$ and a learning rate of $3 \times 10^{-4}$.
 
\subsection{Simulation Results}
 
\begin{figure*}[t]
    \centering
    \includegraphics[width=0.85\textwidth]{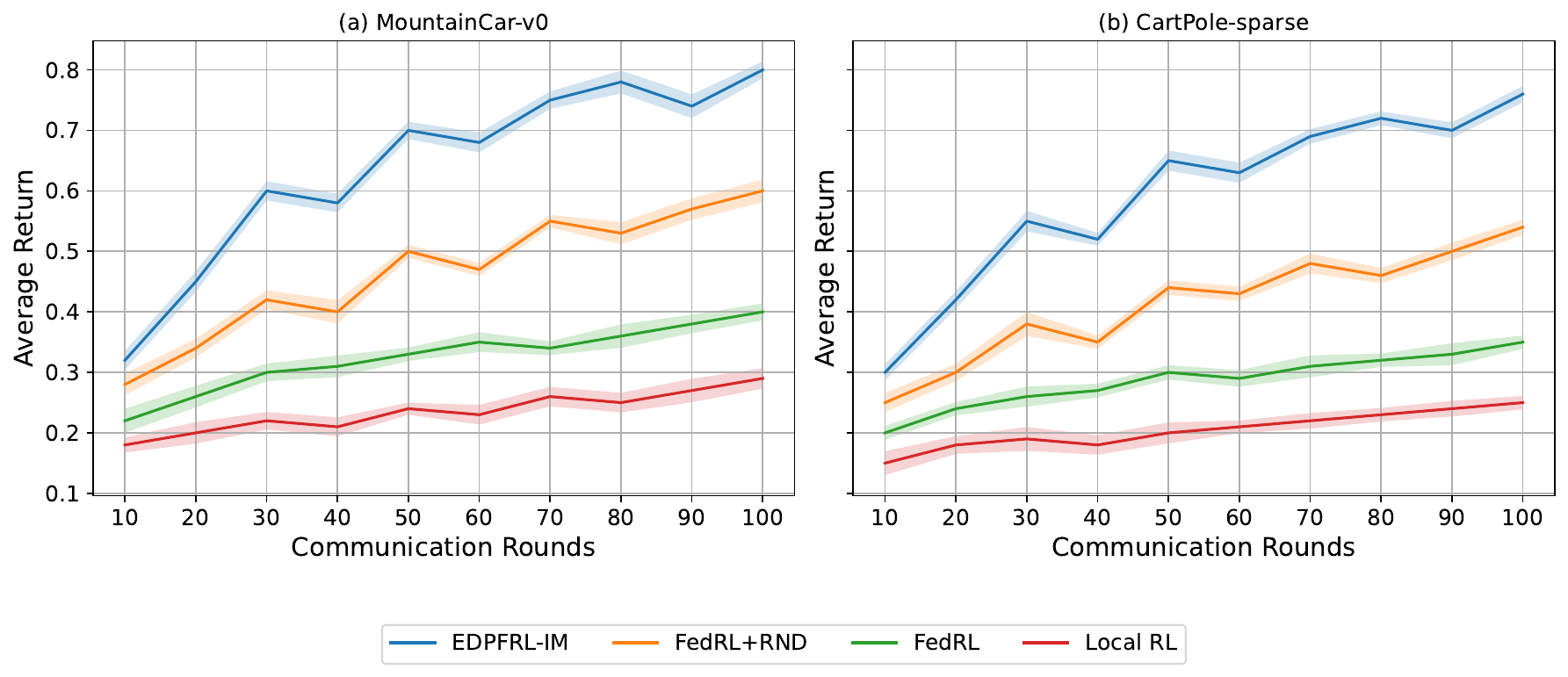}
    \caption{Performance comparison of reinforcement learning methods across two environments over 100 communication rounds. The left subplot shows the average return for \texttt{MountainCar-v0}, and the right subplot shows the average return for \texttt{CartPole-sparse} with four baseline methods.}
    \label{fig:performance_main}
    \vspace{-1em}
\end{figure*}

\textbf{Main Performance Comparison:} Fig.~\ref{fig:performance_main} shows the average return in MountainCar-v0 (left) versus CartPole-sparse (right) across communication rounds. EDPFRL-IM outperformed all baselines (Local RL, FedRL, and FedRL+RND) consistently throughout the initial rounds, with significantly higher returns due to the efficient use of available samples. In MountainCar-v0, EDPFRL-IM achieved quick convergence with a smooth learning curve despite the presence of sparse rewards and difficult state transition dynamics. In CartPole-sparse, EDPFRL-IM had a similar advantage over the other methods. Further, when comparing EDPFRL-IM to FedRL+RND, EDPFRL-IM produced greater average returns and exhibited a more consistent rate of convergence, supporting the claim that coordinated exploration through aggregated novelty priors contributed to superior performance across the range of tested conditions.
 
\textbf{Personalization Benefit:} Table~\ref{tab:client_returns} presents the mean returns for each client from both MountainCar-v0 and CartPole-sparse, with EDPFRL-IM producing greater performance than the FedRL baseline across all clients. The consistent gains highlight the benefit of combining intrinsic motivation with personalization. Additionally, the lower performance variance resulting from EDPFRL-IM allows for greater consistency and personalization of policy learning, which is critical across a wide range of diverse conditions.
 
\begin{table}[t]
\centering
\footnotesize
\caption{\footnotesize Average return per client for EDPFRL-IM and FedRL across two environments. EDPFRL-IM consistently outperforms FedRL in all clients.}
\setlength{\tabcolsep}{2.2pt}
\label{tab:client_returns}
\begin{tabular}{l|cccccccccc}
\toprule
Method & C1 & C2 & C3 & C4 & C5 & C6 & C7 & C8 & C9 & C10 \\
\midrule
\multicolumn{11}{c}{\textbf{MountainCar-v0}} \\
\midrule
EDPFRL-IM & 0.72 & 0.68 & 0.70 & 0.66 & 0.69 & 0.74 & 0.71 & 0.73 & 0.75 & 0.70 \\
FedRL     & 0.42 & 0.39 & 0.40 & 0.38 & 0.41 & 0.44 & 0.40 & 0.42 & 0.43 & 0.41 \\
\midrule
\multicolumn{11}{c}{\textbf{CartPole-sparse}} \\
\midrule
EDPFRL-IM & 0.68 & 0.65 & 0.70 & 0.66 & 0.67 & 0.71 & 0.69 & 0.72 & 0.70 & 0.68 \\
FedRL     & 0.36 & 0.34 & 0.35 & 0.33 & 0.37 & 0.38 & 0.36 & 0.35 & 0.37 & 0.36 \\
\bottomrule
\end{tabular}
\vspace{-1em}
\end{table}
 
\textbf{Ablation: Role of Coordinated Exploration:} Table~\ref{tab:ablation} shows an ablation study that compares the impact of exploration coordination in our proposed strategy. When intrinsic motivation is completely removed, as with the FedRL implementation, agent performance is low due to the lack of exploratory behavior. Adding only the intrinsic reward component (FedRL+RND) provides some improvement but lacks clear directionality (i.e., no consensus on a path). Removing the global novelty prior from EDPFRL-IM (i.e., $\beta = 0$) improves performance slightly through individualized customization; however, performance is still substantially below the full method. EDPFRL-IM achieves the best returns in both environments when coordinated exploratory behavior is used (i.e., $\beta = 0.5$), demonstrating the value of leveraging cross-client novelty to guide exploration in federated reinforcement learning settings.
 
\begin{table}[t]
\centering
\caption{Ablation study: average return over the final 10 communication rounds (rounds 91--100). Coordinated exploration (EDPFRL-IM) leads to the best performance across environments.}
\label{tab:ablation}
\begin{tabular}{l|c|c}
\toprule
\textbf{Method Variant} & \textbf{MountainCar-v0} & \textbf{CartPole-sparse} \\
\midrule
FedRL (no RND)               & 0.40 & 0.35 \\
FedRL+RND (no coordination) & 0.60 & 0.54 \\
EDPFRL-IM ($\beta$ = 0)             & 0.63 & 0.58 \\
EDPFRL-IM (ours)              & \textbf{0.80} & \textbf{0.76} \\
\bottomrule
\end{tabular}
\vspace{-1em}
\end{table}
 
\begin{figure*}[t]
    \centering
    \includegraphics[width=0.85\textwidth]{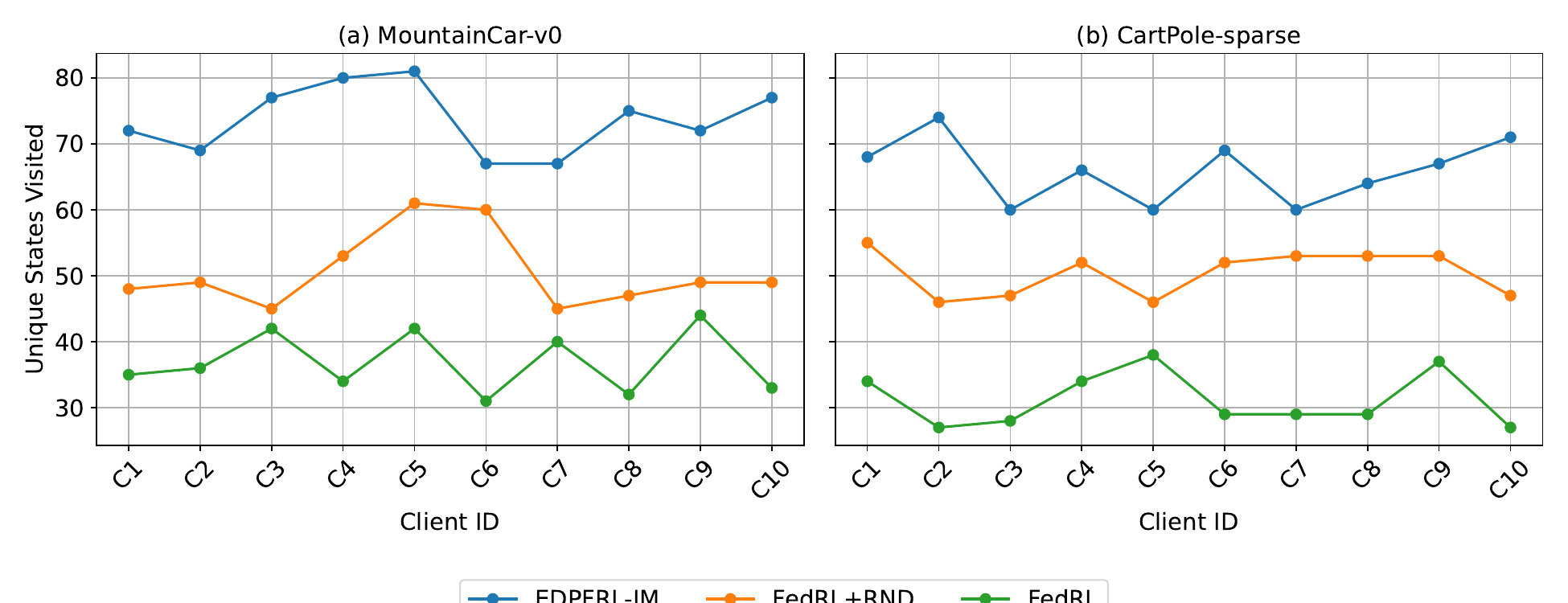}
    \caption{
    Exploration coverage across clients in two environments.
    EDPFRL-IM enables broader state-space exploration by coordinating curiosity across clients.
    }
    \vspace{-1em}
    \label{fig:exploration_coverage}
\end{figure*}

\textbf{Exploration Coverage Analysis:} As illustrated in Fig.~\ref{fig:exploration_coverage}, clients' exploration coverage (percentage of the state space that clients visit) in both environments is much greater when using EDPFRL-IM compared to using FedRL and FedRL+RND. This indicates that the coordinated use of intrinsic motivation between clients is crucial for increasing the diversity of exploration. Increased diversity in exploration provides for greater generalization and faster adaptation in heterogeneous environments.
 
\begin{figure*}[t]
    \centering
    \includegraphics[width=0.85\textwidth]{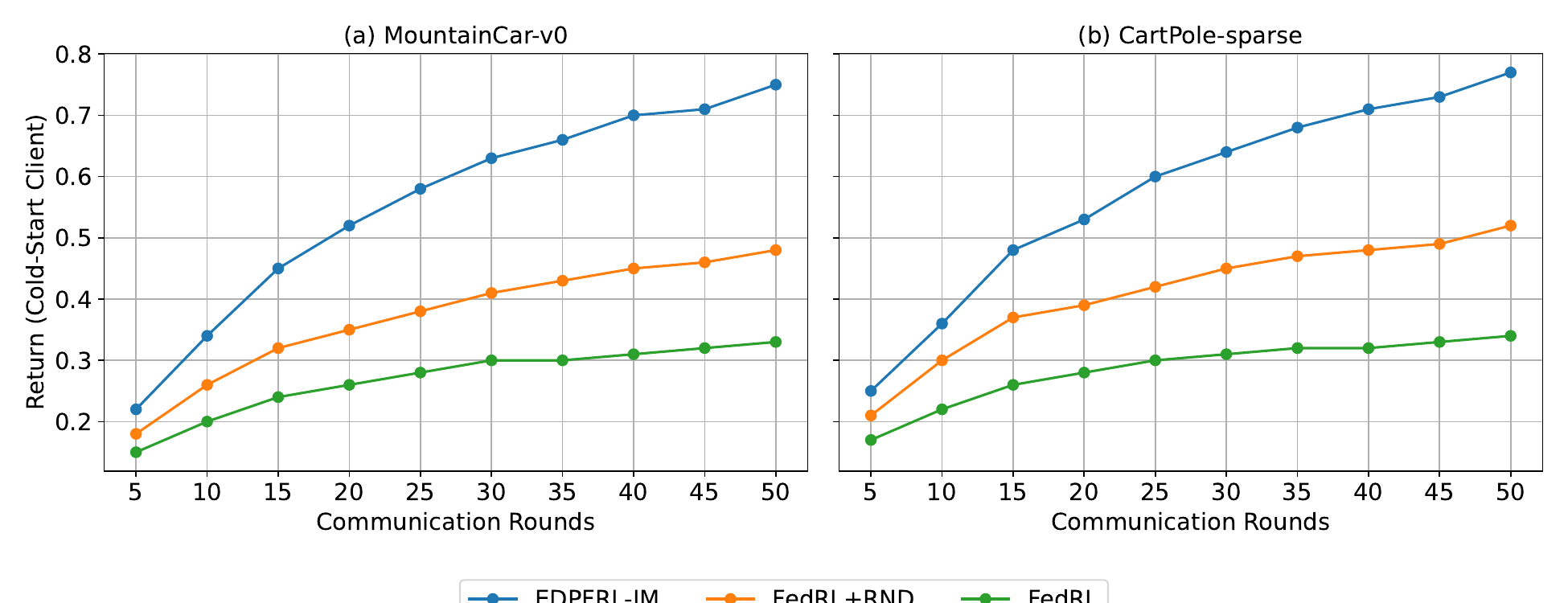}
    \caption{
    Return trajectories of a cold-start client introduced at round 5 in (a) \texttt{MountainCar-v0} and (b) \texttt{CartPole-sparse}. EDPFRL-IM quickly adapts by leveraging the global exploration prior, outperforming both FedRL and FedRL+RND.
    }
    \label{fig:cold_start_combined}
    \vspace{-1em}
\end{figure*}

\textbf{Robustness to Cold Start Clients:} We assessed the adaptability of the three methods to the introduction of a cold-start client during training (at round 5) by measuring how quickly and how well the cold-start client performed in both environments, as shown in Fig.~\ref{fig:cold_start_combined}. EDPFRL-IM uses globally aggregated exploration knowledge to facilitate rapid and smooth adaptation for cold-start clients in both environments, whereas FedRL and FedRL+RND do not use this form of globally aggregated knowledge, which negatively affects their ability to support early-stage learners, resulting in slower progress.
 
\textbf{Comparison with Existing Approaches:} Table~\ref{tab:final_comparison} presents the final average return achieved by several federated and personalized reinforcement learning algorithms. EDPFRL-IM outperformed all comparison methods, including RND, ICM, CBE, FedAvg-RL, pFedMe, FedPer++, and FedRL+RND, in both environments. Although approaches such as pFedMe and FedPer++ support personalization of RL models, they provide limited exploration capabilities for discovering new behaviors. FedRL+RND incorporates intrinsic motivation through curiosity-driven exploration but lacks an efficient mechanism for personalized federated adaptation across clients. The proposed EDPFRL-IM framework combines structured curiosity-driven exploration with federated personalization, yielding significantly improved performance and learning efficiency compared with existing methods.
 
\begin{table}[t]
\centering
\footnotesize
\caption{Final Average Return Comparison Across Methods}
\label{tab:final_comparison}
\begin{tabular}{lcc}
\toprule
\textbf{Method} & \textbf{MountainCar-v0} & \textbf{CartPole-sparse} \\
\midrule
RND \cite{nikulin2023anti}         & 0.31 & 0.32 \\
ICM \cite{colas2019curious} & 0.34 & 0.36 \\
CBE \cite{machado2020count}   & 0.39 & 0.49 \\
FedAvg-RL \cite{wang2020optimizing}        & 0.38 & 0.42 \\
pFedMe \cite{t2020personalized}       & 0.40 & 0.45 \\
FedPer++ \cite{xu2022fedper++}      & 0.41 & 0.44 \\
FedRL+RND \cite{fan2021fault,nikulin2023anti}    & 0.48 & 0.52 \\
\textbf{EDPFRL-IM (Ours)} & \textbf{0.76} & \textbf{0.74} \\
\bottomrule
\end{tabular}
\vspace{-1em}
\end{table}
 
\section{Conclusion}
 
This paper presents EDPFRL-IM, an exploration-driven personalized federated reinforcement learning (PFRL) system that leverages intrinsic motivation to provide coordinated yet private client exploration. EDPFRL-IM addresses an important shortcoming of prior, less effective PFRL systems by incorporating a global novelty prior computed from a compressed set of client exploration statistics, enabling each client to learn while adapting to its local context and benefiting from the collective experience of the remaining clients. Comprehensive evaluation of EDPFRL-IM in environments with sparse reward structures and high variability demonstrates that it achieves significant improvements in sample efficiency and personalization, and adapts faster for cold-start clients than its peers. EDPFRL-IM consistently outperforms baseline systems using both standard and personalized FL algorithms, including various intrinsic-incentive-based FL algorithms, and does so with no compromise in terms of privacy.

\bibliographystyle{ieeetr}
\bibliography{ref}

\end{document}